# A survey of recent methods for addressing AI fairness and bias in biomedicine

Yifan Yang[1,2], Mingquan Lin[3], Han Zhao[4], Yifan Peng[3], Furong Huang[2], Zhiyong Lu[1,*]

[1]National Center for Biotechnology Information (NCBI), National Library of Medicine (NLM), National Institutes of Health (NIH), Bethesda, MD, USA

[2]Department of Computer Science, University of Maryland, College Park USA

[3]Department of Population Health Sciences, Weill Cornell Medicine, New York, USA

[4]Department of Computer Science, University of Illinois at Urbana-Champaign, Champaign, IL

*Corresponding author

Zhiyong Lu, PhD FACMI FIAHSI

zhiyong.lu@nih.gov; 301-435-4920

**ABSTRACT**

**Objectives:** Artificial intelligence (AI) systems have the potential to revolutionize clinical practices, including improving diagnostic accuracy and surgical decision-making, while also reducing costs and manpower. However, it is important to recognize that these systems may perpetuate social inequities or demonstrate biases, such as those based on race or gender. Such biases can occur before, during, or after the development of AI models, making it critical to understand and address potential biases to enable the accurate and reliable application of AI models in clinical settings. To mitigate bias concerns during model development, we surveyed recent publications on different debiasing methods in the fields of biomedical natural language processing (NLP) or computer vision (CV). Then we discussed the methods, such as data perturbation and adversarial learning, that have been applied in the biomedical domain to address bias.

**Methods**: We performed our literature search on PubMed, ACM digital library, and IEEE Xplore of relevant articles published between January 2018 and December 2023 using multiple combinations of keywords. We then filtered the result of 10,041 articles automatically with loose constraints, and manually inspected the abstracts of the remaining 890 articles to identify the 55 articles included in this review. Additional articles in the references are also included in this review. We discuss each method and compare its strengths and weaknesses. Finally, we review other potential methods from the general domain that could be applied to biomedicine to address bias and improve fairness.

**Results:** The bias of AIs in biomedicine can originate from multiple sources such as insufficient data, sampling bias and the use of health-irrelevant features or race-adjusted algorithms. Existing debiasing methods that focus on algorithms can be categorized into distributional or algorithmic. Distributional methods include data augmentation, data perturbation, data reweighting methods, and federated learning. Algorithmic approaches include unsupervised representation learning, adversarial learning, disentangled representation learning, loss-based methods and causality-based methods.

## 1. INTRODUCTION

Over the past decade, Artificial intelligence (AI) systems that use deep neural networks have become increasingly popular in medicine. AI's diagnostic capabilities have proven to be on par with domain experts in multiple specialties, such as pathology classification, anatomical segmentation, lesion delineation, cardiovascular medicine, diabetic retinopathy, skin cancer, pneumonia, and hepatocellular cancer.[1–6] For example, AI models fed with live streaming Electronic Health Record (EHR) data in surgical decision-making can address the weakness of time-consuming manual data entry and suboptimal accuracy.[7] In ophthalmology, the application of AI models can reduce the usage of manpower and the cost of screening, including in remote settings.[4,8] Using acoustics, MRIs, CTs, or x-rays, AI systems can screen for indicators of osteoporosis, such as bone mass or fragility fractures, or automatically segment images of patients with or at risk of osteoporosis.[9]

Despite the remarkable achievements of AI, there are still substantial concerns regarding the fairness and bias of AI models in the field of biomedicine. Prior work defines bias in healthcare as systematic error due to flaws in the study's design, conduct, or analysis.[10] This type of error can bias the study results and make it difficult to accurately interpret the findings. The issue of bias and fairness in AI systems has been the subject of extensive research and discussion in the broader field of AI and computer science for many years.[11–13] In this work, the emphasis is on biases related to demographic or geographic subgroups. We define bias as the presence of systematic errors or disparities within decision-making processes that disproportionately affect specific subgroups. Conversely, fairness is defined as the eradication/diminishing of these biases to ensure outcomes that are both equitable and just. In machine learning, there are many different fairness definitions, focusing on different aspects of the evaluation.[14] For example, demographic parity requires the probability of positive prediction to be the same across different sub-groups, and accuracy parity focuses on equalizing the error rate of sub-groups. In the context of biomedicine, the criteria for making subgroups can be demographic information such as race, sex, or age, and can also be things like admitted hospitals and brand of radiographic devices used in diagnosis.

The unequal behavior of algorithms toward different population sub-groups may be considered a violation of the principles of bioethics, which include justice, autonomy, beneficence, and non-maleficence.[3] When predicting in-hospital mortality, physiological decompensation, length of stay, and phenotype classification, models have demonstrated lower performance for minority populations.[15] Studies have shown that AI models can predict unnecessarily protected information, such as gender, race, and institution, from the representation of scan images.[16–18]

Several reviews in the biomedical domain have summarized the issue of fairness in artificial intelligence[3,19–21]. Unlike these studies, we focus on existing debiasing methods from biomedical natural language processing (NLP) or computer vision (CV). We discuss each method, compare their strengths and weaknesses, and examine what problem each category of method can potentially solve. Finally, we discuss other potential methods from the general domain that could be applied to biomedicine to address bias and improve fairness, and provide future directions on fairness in biomedicine AI.

| Problem | Artificial Intelligence (AI) systems perpetuate social inequities or demonstrate biases, such as those based on race or gender. |
|---|---|
| What is Already Known | AI systems have the potential to revolutionize clinical practices, including improving diagnostic accuracy and surgical decision-making, while also reducing costs and manpower. It is critical to understand and address potential biases to enable the application of AI models accurately and reliably in clinical settings. |
| What This Paper Adds | We surveyed recent papers on AI bias and then discussed the source of bias for biomedicine AI. We also include methods, such as data perturbation and adversarial learning, that have been applied in the biomedical domain to address bias. We also discuss methods that have been applied in the general domain to address bias that could be applied to clinical tasks. Understanding the Taxonomy of methods and the source of bias would help choose the correct measure to address bias in various situations. |

## 2. MATERIALS AND METHODS

We performed our literature search on PubMed, ACM digital library, and IEEE Xplore. The overall pipeline is demonstrated in Figure 1. We searched peer-reviewed articles on PubMed between January 2018 and December 2022 with the combination of two sets of keywords: *(AI or ML or Deep learning or Algorithmic) + (equity or bias)*. On the ACM digital library and IEEE Xplore with the same time frame, we included

"medical" in addition to the above constraint. The total number of articles was 10,041. After removing duplicates, we filtered the abstracts of these results with additional constraints on the PubMed articles, resulting in 890 articles. The additional constraint on PubMed articles checks whether the term "bias" in the article only comes from the phrase "Risk of bias", a commonly used term in reporting medical studies. We then manually screened abstracts of these articles and found 55 related to using deep learning in biomedicine to address equity issues. We included relevant articles in their references. Articles that advocated for more research or presented the current state of the problem overlapped were excluded. The search was supplemented with a general internet search of debiasing methods, and a search focused on key debiasing methods in the general domain.

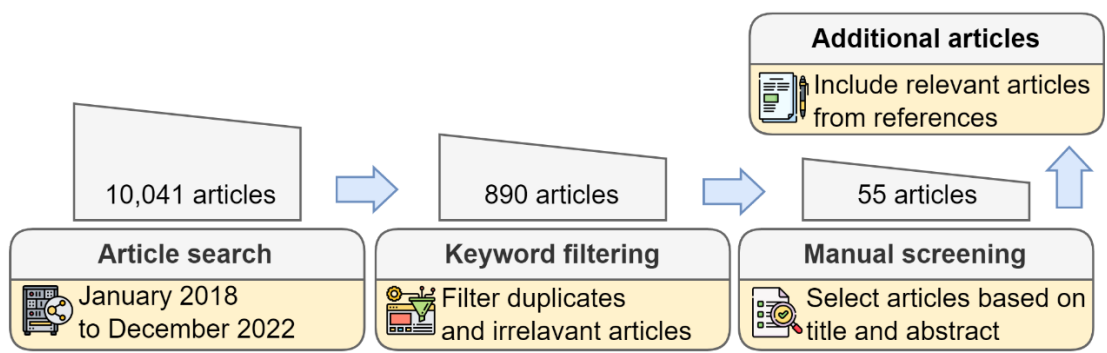

**Figure 1.** Pipeline of searching and filtering the literature for this review

## 3. RESULTS

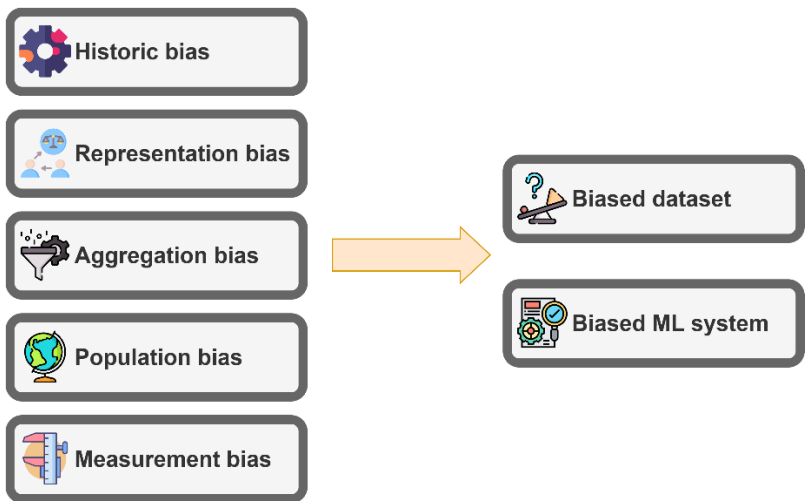

**Figure 2.** Different sources of bias that can lead to biased datasets and ML systems in biomedicine

### 3.1 Causes of Biases

To apply AI models accurately and reliably to clinical settings, it is necessary to understand and address potential biases. Bias can occur at any stage of the process, from data collection to model development[22]. Mehrabi et al. present a categorization of bias in ML systems, delineating them into three primary types: biases originating from data to algorithm, from user interaction to data, and from algorithm to user interaction[23]. Additionally, they elaborate on 19 distinct types of biases within these categories. Similarly, Suresh et al. pinpointed 7 sources of bias in the ML system life cycle that rages from data collection to model deployment[24]. While these taxonomies are broadly applicable in ML fairness, the biomedicine domain faces unique challenges and specific bias sources. Biases in biomedicine AI not only align with general AI biases but also include distinct causes pertinent to this field, such as disparities in healthcare access, disease prevalence variations across populations, and differing quality of medical data.

This context specificity is crucial for understanding and mitigating biases in biomedical AI. Figure 2 illustrates the various sources of biases specifically in biomedical AI systems. These biases significantly impact ML system outcomes, potentially favoring or disadvantaging specific demographic or geographic subgroups, or leading to biased data collection that can result in unfair ML systems with serious implications for healthcare equality and efficacy.

### 3.1.1 Historic bias

Historical bias, stemming from existing societal and technological issues, can exist in the data generation processes, established tools or pretrained models[23]. A 2019 study revealed historic bias related to race in heart-failure management at a Boston emergency department, influenced by the American Heart Association's (AHA) scoring system. This system, the Get with the Guidelines–Heart Failure Risk Score, assigns higher risk points to "nonblack" patients, resulting in black and Latinx patients being less likely than white patients to be admitted to cardiology services[25]. In a study by Zhang et al., they used MIMIC-III[26], a publicly available clinical note database, to finetune a BERT model initialized from SciBERT[27]. The researchers found that modifying race can generate a worse course of action for African American patients and observed a prediction gap between different genders and races, suggesting the inherent bias of pretrained models[28]. The underlying disadvantage for African Americans learned by the language model is a type of historic bias.

### 3.1.2 Representation bias

Representation bias arises when a development sample does not adequately represent a section of the population, leading to its inability to effectively generalize for a specific subset of the user population[24]. For instance, in diabetic retinopathy, the data imbalance results in a large accuracy gap between light-skinned and dark-skinned subjects[3]. Another example is facial recognition software with AI for early detection of acromegaly, where existing studies have been performed only in predominantly White or Asian populations[29]. Considering that the data collected is always biased towards the population that can afford healthcare, the representation bias in biomedicine AI is also related to the historic bias[2,21,29–32].

### 3.1.3 Aggregation bias

Aggregation bias occurs when incorrect assumptions about individuals or sub-groups are made based on observations of the entire population[23]. This can happen when making predictions about a subgroup based on evidence from a larger population. For example, in precision medicine for selection for type 2 diabetes, a large patient group better suited for SGLT2 inhibitor therapy was under-recommended for it due to low overall cardiorenal disease risk, with only 14.3% receiving the suggestion. Conversely, a smaller group, ideal for DPP4 inhibitor therapy, had 75% recommended SGLT2 inhibitors instead, influenced by their older age and higher cardiorenal risk[31].

### 3.1.4 Population bias

Population bias in biomedical AI emerges when the characteristics, demographics, and statistical profiles of the patient population differ significantly from those of the intended patient population[24]. For example, predictive models trained on data dominated by White Americans result in higher error rates for African Americans[33]. In the context of cardiac imaging technologies and digital histology, the quality of images produced by cardiac MRI machines and the coloring and quality of uploaded tumor tissue slides can vary among institutions and vendors, resulting in site-specific signatures or conventions that can introduce population bias when applying trained models to different cites[2,8,18].

### 3.1.5 Measurement bias

Measurement bias occurs due to the selection, usage, and measurement of specific features[23]. An algorithm using healthcare costs may incorrectly conclude that dark-skinned patients are healthier than light-skinned patients with equal levels of illness, as they may receive less healthcare funding. This may result in a significant reduction in the number of dark-skinned patients receiving treatment[22,25], and healthcare costs can be viewed as a mismeasured proxy.

In this review, we focus on methods that can be applied during model development. We adopt a modified taxonomy from a recent work[34] and categorize the recent debiasing methods into two groups: distributional and algorithmic (Table 1). Distributional debiasing methods aim to change the underlying data distribution of the dataset to improve fairness, whereas algorithmic debiasing methods modify the training procedure or model architecture to ensure fairness. We excluded methods that collect extra data through crowdsourcing or other means of human annotation, as these approaches are usually more expensive or difficult to obtain in the biomedical domain.

| Distributional methods | *Data augmentation [35–45] | Increase size and diversity by generating synthetic data and can help anonymize source data to increase availability. |
| | Data perturbation [46–49] | Increases the dataset diversity by altering demographic information on |

| | | existing data. Mostly applicable to text data. |
|---|---|---|
| | *Data reweighting [50,51] | Compensate under-represented subgroups by duplicating those samples. |
| | *Federated learning [52,53] | Allow the central model to be exposed to data from various sources by merging training results from multiple centers. |
| Algorithmic methods | *Unsupervised representation learning [54–58] | Can be used to learn models that extract useful features with unlabeled small datasets. |
| | *Adversarial learning [59–63] | Removes bias by training the model to forget protected attributes. |
| | Disentangled representation learning [64–67] | Disentangles the learned representation into protected and target attributes, and promotes fairness and explainability by only using the target attribute. |
| | *Loss function [67–69] | Optimize the model directly to achieve fairness or equivalent constraint. |
| | Causality [70,71] | Identifies stable data relationships across various contexts to build models resilient to input changes and biases |

**Table 1.** List of debiasing model-driven methods for AI systems in CV, NLP, and biomedicine covered in this work

*Note.* * denotes that the method has been applied to address equity in the biomedicine domain.

## 3.2 Distributional methods in the biomedical domain and general domain

### 3.2.1 Data augmentation

One approach to reducing bias in machine learning models is to balance the dataset, as training models with balanced and high-quality datasets may result in reduced bias. There are two ways to balance a dataset: remove data from dominant groups or add data in the under-represented group. Collecting new data, however, is often time-consuming and costly. Hence, a common approach is to create synthetic data in the under-represented group using algorithms such as generative adversarial networks (GANs). This approach is also commonly referred to as *data augmentation*.

GANs are machine learning models that generate synthetic samples similar to a given input dataset.[35] GANs typically consist of two main components: a generator model and a discriminator model. The generator model aims to generate samples similar to real data, while the discriminator model attempts to distinguish between real and generated samples. The two models are typically trained simultaneously, with the generator model attempting to fool the discriminator model and the discriminator model trying to correctly identify generated samples. This adversarial training process can lead to the generation of high-quality synthetic samples. There are many applications of GANs in ophthalmology, such as the generation of the synthetic fundus, optical coherence tomography (OCT), or fluorescein angiography images.[36]

Studies have shown success in augmenting the medical dataset with GANs to address the bias in the dataset. Burlina et al. used GANs to change the retinal pigmentation in retinal fundus images and create a more balanced dataset.[37] Coyner et al. demonstrate that PGAN, a method that progressively increases the size of GAN models during training to stably produce large and high-quality images, is able to synthesize retinal vessel maps that better represent the distribution of plus disease compared to the original dataset, potentially reducing bias due to limited sample size.[38] Conditional GANs allow infusion of external information when generating images, which can help counterbalance biased datasets by generating data in the less represented subgroups.[39] In addition to addressing issues with data imbalance and sparsity, the images generated by GANs are distinguishable from the original dataset, potentially reducing privacy concerns associated with sharing the generated data and enabling further research. GANs are also widely applied in CV as a debiasing data augmentation method.[40,41] Diffusion models are also gaining popularity. Although such models have not been widely applied to debiasing, the nature of such generative models makes them similar to how GANs are used in debiasing. Diffusion models learn a function that recovers data from noise. In the training process, data such as images are first transformed into Gaussian, then back to images from Gaussian.[42] The diffusion steps can be treated as a long Markov chain.

Prior work has also shown that using a text-guided prompt or a classifier for conditional image synthesis, which can be applied in debiasing a dataset, is promising.[43,44] A recent work, RoentGen, explores the potential of latent diffusion models on Chest X-ray image generation.[45] RoentGen is able to use a text prompt as a guide to synthesizing X-ray images and demonstrates that combining synthesized and real data achieves the highest accuracy. This is promising, as the method also can address dataset equity problems in the biomedicine domain.

### 3.2.2 Data perturbation

Data perturbation increases the diversity and balance of the dataset by adding "noise" to existing samples. Under the concept of data perturbation, CDA is a dataset debiasing method that involves adding crafted samples, usually through a template or heuristic.[46,47] In NLP research, CDA is used to craft samples by replacing words that may cause bias with words that refer to under-represented groups. For instance, to reduce gender bias in training data, CDA would modify the phrase, "He is a doctor," to "She is a doctor," to generate new samples in the dataset.[46] Addressing the limitations of heuristic-based generation, a recent work makes use of a sequence-to-sequence model to generate perturbated sentences of high quality.[48] The work of Burlina et al. can be viewed as a combination of CDA and GANs, whereby images of another protected attribute were generated through GANs.[37] Similar to CDA, Minot et al. propose to remove gender-specific words in the dataset, and show that their model can achieve a fairer performance using the truncated HER on health-condition classifications[49].

### 3.2.3 Data reweighting

A surprisingly popular approach is to duplicate the minority class data. Acknowledging that word embeddings generated from language models can be biased, Agmon et al. duplicate clinical trials that include more female participants in the training data and gain a slight improvement in the female subclass in comorbidity classification, hospital length of stay and ICU readmission prediction.[50] Similarly, Afrose et al. propose a double-prioritized bias correction that replicates the minority demographic subgroup to address data imbalance in mortality prediction.[51]

### 3.2.4 Federated Learning

Federated learning is a class of methods that uses multiple devices, often distant from each other, to train models. In federated learning, a central machine aggregates learning from other devices that are often referred to as clients. It has the benefit of preserving privacy while utilizing multiple data sources, addressing the data issue. A common source of bias in biomedicine comes from limited data sources or skewed demographic distribution in the dataset. Federated learning allows data usage of various centers, potentially exposing the model to different demographic populations. One key to federated learning algorithms is aggregating weights returned from different devices. Using multiple fairness aggregation algorithms, Meerza et al. show that it is possible to learn a fair prediction on Alzheimer's Disease using spontaneous speech data from daily collections[52]. FEderated LearnIng with a CentralIzed Adversary (FELICIA) is another method based on federated learning, and has shown effectiveness on skin lesion classification[53]. FELICIA makes the adversarial discriminator component of GAN as the central control, and the communication between the central discriminator and the clients only consists of synthetic images, preserving privacy while achieving fairness.

### 3.2.5 Discussion on distributional methods

Distributional methods offer solutions to the challenge of insufficient non-diverse data. Prior research has shown that techniques such as synthesizing dark-color retinal images or collaborating with different centers through federated learning can effectively address the issue of this challenge. Moreover, in the context of biomedicine NLP tasks, such as search retrieval and report generation, data perturbation techniques can also yield potential benefits.

GAN models have been widely studied and shown to have state-of-the-art results. However, studies have shown that they capture less diversity than other models, such as diffusion-based ones. It is often difficult and time-consuming to train GAN models, and they have a collapsing problem, whereby the generator model produces a few images.[43] In contrast, diffusion models are faster and more stable in training. Due to the long Markov chain process, however, the generation time is longer than in GAN models. The generated images may also contain existing biases in the training data, introduce new biases, and training on generated images may make the model worse[72]. Counterfactual data augmentation (CDA) methods, in comparison, are mostly rule-based; however, they are applicable to only text data. Duplicating data in the minority subgroups requires a lot of fine-tuning and would take a lot of time. An easier operation would be to implement different weights to samples. Both Agmon et al. and Afrose et al. explore different configurations of the number of times that minority data are duplicated, and the improvements are limited.[50,51] .

While federated learning increases data diversity, applying this method in reality is often burdensome. Different centers may have different data formats and computation power. The communication between the central machine and clients is another bottleneck of applying federated learning.

## 3.3 Algorithmic methods in the biomedical domain and general domain

### 3.3.1 Unsupervised learning

Bias can occur when the dataset is small, such as under-represented groups in a large dataset or datasets of rare diseases. Recent work proposes to address biases in the model by using Deep InfoMax or its variant, Augmented Multiscale Deep InfoMax (AMDIM).[54,55] These architectures are designed to learn image representations in a self-supervised manner, using techniques such as predicting part of an image using the whole image, like a cloze task in computer vision. This self-supervised learning approach allows the model to learn useful representations of the data without the need for external labels. The goal of Deep InfoMAX and AMDIM is to generate unique, view-invariant representations of images that also maintain local consistency across structural locations within the image.[54,55] Burlina et al. find that using the representation learned by AMDIM, the performance of downstream tasks (such as disease prediction) was acceptable even when using relatively small training sizes; in addition, the performance degradation when using very small training sizes was less severe.[56]

Variational Auto-Encoder (VAE) is another commonly used unsupervised representation learning algorithm. This method learns by decomposing and reconstructing the input. Building from VAE, Conditional VAE (CVAE)[57] introduces a variational posterior as an approximation of the true posterior. Using Conditional Variational Auto-encoder, Liu et al. propose a method to learn a fair prior distribution of the dataset, and show reduced bias in EHR modeling [58].

### 3.3.2 Adversarial learning

Protected attributes, such as race, should not be used in algorithms as a feature.[17] One way to debias AI models is to force them to not learn the protected attributes. Hard debiasing is a technique that reduces bias in a model's representations by training a linear classifier to predict the protected attributes. The technique propagates the sign-flipped gradient of the classifier that predicts the protected attributes. Through such adversarial training, the model avoids learning protected attributes, such as institution and gender,[28] or forgets the learned protected attributes. Prior work in colorectal cancer prediction has shown that after debiasing, the representation space is more invariant to the bias variables. In contrast, the baseline representation space shows a clear correlation between the protected attributes and the learned representation.[59] There has also been success in reducing the effect of the biased features of deep learning models on histopathology images through adversarial training.[60] Other than adversarial training or finetuning the whole model, Wu et al. propose FairPrune, a method to promote fairness by determining and pruning the subset of a model that is contributing to bias,[61] and show that it is effective in dermatology disease classification. The authors use the second derivative of model parameters to quantify their importance and remove a certain percentage of parameters repeatedly until the target fairness metric is achieved.

Instead of using a sign-flipped gradient in adversarial learning, other methods, such as INP, can be used to remove the unwanted information learned by the classifier.[62] This is done by iteratively projecting the representation that contains protected information into the null space of the classifier's weight matrix.

Following the philosophy of adversarial learning, Zanna et al. hypothesis that when the model is the most uncertain about protected labels, the weights of the model can be fully utilized for the task[63]. By estimating the model's uncertainty, they propose a bias mitigation technique for anxiety prediction that also preserves fairness.

### 3.3.3 Disentangled representation learning

A type of method related to content-style disentanglement is disentangled representation learning. Representations learned by AI systems are information or features extracted from data. Such representation is a combination of many features.

Disentangling representations into subspaces and excluding the subspace related to protected attributes can promote a fairer AI system. It is worth noting that disentangled representation learning has been used in the biomedicine domain to address problems such as disease decomposition, artifact reduction, harmonization, and so forth.[73] Research, however, focuses mainly on methods that disentangle the style and content of images. Disentangled representation learning might be more effective (in terms of accuracy) in a low-resource setting; however, the improvement is limited.[64] A prior work modified the vanilla variational autoencoder to produce flexibly fair representations that can be applicable in many datasets.[65] Based on the concern that removing all protected attributes might reduce the model's performance, a common situation in the biomedical setting, others have proposed subcategorizing a mutual attribute latent space in addition to the target and protected attribute latent representation space, which allows for less loss of information.[66]

### 3.3.4 Loss function

Machine learning algorithms are trained to minimize their loss function. Thus, infusing a fairness constraint in the loss function can help learn a fair model. Using Rènyi's divergence, Gronowski et al. derive a loss function via a variational approach to promote fairness in diabetic retinopathy detection[67]. Serna et al. propose a discrimination-aware loss function for face recognition algorithms based on a triplet loss function and a sensitive triplet generator[68]. Zafar et al. add a trackable constraint to limit model's decision boundaries and unfairness.  By adding a fairness constraint to ensure similar individuals are treated similarly, Dwork et al.'s method guarantees statistical parity[69].

### 3.3.5 Causality

Causal invariant learning, a technique that identifies stable data relationships across varied environments, helps build models resistant to input biases and changes. A common theorem applied in causal invariant learning is backdoor adjustment[74], which aims to eliminate the effect of spurious correlations that is often the reason for biased models. Following this theorem, Causal Intervention by Semantic Smoothing (CISS) learns the causal effect between text input and labels for robust predictions[70]. Causal Intervention by Instrumental Variable (CiiV) employs retinotopic sampling masks and consistency regularization loss, encouraging neural networks to prioritize causal features over local confounders, thereby boosting adversarial robustness in visual tasks[71].

### 3.3.6 Discussion on algorithmic methods

Algorithmic methods inherently address bias issues from the model's perspective, offering a valuable approach to mitigating bias, especially in cases where data collection exhibits inherent bias. These methods serve to safeguard the model against the inadvertent exploitation of undesired distributional information, thereby promoting fairness and reducing the potential for bias-related issues.

Unsupervised representation learning with fill-in-the-blank tasks, such as a cloze task, is common in masked language model pretraining, such as BERT,[75] and it has been shown that this type of pretraining significantly improves convergence speed and performance for downstream tasks.[76] The results from Burlina et al. show, however, that the improvement is not significant when the dataset is large, suggesting that the method improves performance only when the dataset is small.

Adversarial methods/algorithms that aim to remove sensitive information are difficult to train and are not so effective, as classifiers are still able to extract the protected attributes. Additional training after debiasing might improve the model's performance[77]; however, whether this will make the model regain the ability to learn protected attributes is unclear. Iterative null space projection (INP) does not guarantee the removal of desired attributes and may not work when the representation is fed into a non-linear classifier.

Loss function can be very sensitive to model architecture and regularization parameters. A loss function may work on one type of data but fail to converge on another.

A key limitation of causality-based methods lies in their dependency on the accuracy of the presumed causal relationships. Misidentifying or overlooking these relationships can significantly impact their effectiveness. For example, incorrectly assuming no causal relationships between breast cancer and sexuality will lead to inaccurate and ineffective methods.

## 3.4 Beyond model-oriented methods

While the primary emphasis of this study lies in model-oriented approaches, we also wish to underscore the significance of endeavors extending beyond these methodologies. The literature we reviewed discussed methods that are not model oriented to address biases and fairness in AI systems applied in biomedicine. Indeed, the workflow of AI models, especially in the biomedical domain, involves not just model development but also data collection, system deployment, system operation, and so forth.

In addition to the model-oriented methods of addressing bias in AI systems, there are important aspects to consider, one of which is explainability or transparency, the ability of the algorithm to provide clear and understandable reasoning for its predictions or decisions. The lack of transparency in AI systems can make it difficult to determine whether the system is biased or to identify ways to address any existing biases.[21] Improving transparency is also key to advancing the use of AI systems in clinical settings[5,7,21,30,78]; it can help to ensure that the algorithms are being used ethically and in a way that aligns with the goals of the healthcare system. Models such as recurrent neural networks and attention models can help to locate which part of a longitudinal EHR dataset is playing an important role in prediction, whereas saliency maps and SHapely Additive exPlanations (SHAP) have been utilized to identify which area of an image the model attends to in various tasks.[22,59,78,79] Prior research has called for using fairness audits as a common practice instead of as repairments to the flaws of existing work.[3] Protocols such as Prediction model Risk Of Bias ASsessment (PROBAST) and PROBAST-AI help researchers, clinicians, systematic reviewers, and policymakers to evaluate machine learning-based prediction model studies with standardized approaches to bias evaluation, allowing users to critically assess the quality of these studies and make more informed decisions about their utility.[13,80] Data statements that include the curation rationale and demographic characteristics of the dataset should also be included when datasets are published.[81] We list the potential datasets that researchers can use to conduct fairness research in biomedical tasks in Table 2, and more datasets in the medical imaging domain can be found in the work of Xu et al. and Lara et al[3,19]. We strongly recommend that future datasets be enriched by including anonymized demographic information, as this would allow for more nuanced and comprehensive analysis of the methods, and would help to ensure that research and analysis are conducted in a fair and ethical manner.

| Name | Modality | Gender | Age | Race/Ethnicity |
|------|----------|--------|-----|----------------|
| CheXpert[82] | Chest radiograph | √ | √ | √ |
| MIMIC Chest X-Ray[83] | Chest radiograph with report | √ | √ | √ |
| ChestX-ray8[84] | Chest radiograph | √ | √ | X |
| UK Biobank[85] | Cardiac MRI | √ | √ | √ |
| RadFusion[86] | EHR and CT scans for pulmonary embolisms | √ | √ | √ |
| PaPILa[87] | Retinal fundus image and related medical data for glaucoma assessment | √ | √ | X |
| HAM10000[88] | Dermatoscopic images of common pigmented skin lesions | √ | √ | X |
| EchoNet-Pediatric[89] | Pediatric echocardiography | √ | √ | X |
| COVID-CT-MD[90] | CT scans for COVID-19 | √ | √ | X |
| ISIC 2020[91] | Dermatology images for skin lesions | √ | √ | X |

**Table 2.** Publicly available fairness research datasets that contain demographic information.

Such information could include age, gender, ethnicity, and other relevant factors.

It is also important to increase the diversity of datasets. The majority of the datasets used for training machine learning algorithms are sourced from high-income countries in Europe and North America, which introduces a bias, as the demographics of these countries do not accurately represent those of other regions, such as Africa,

Asia, and Latin America.[3] As a result, the AI systems developed using these datasets may not be applicable or effective in other parts of the world. Existing methods for addressing bias due to imbalanced demographic distributions in datasets can help improve the fairness of AI systems. Increasing the diversity of groups represented in the dataset can further promote fairness and improve the performance of the models.

## 4. DISCUSSION

In this work, we identify five major types of sources causing bias and unfairness in medical AI systems. Accurately pinpointing the specific source of this bias is crucial as it informs the choice of the most effective approach to counteract and diminish these biases. For instance, algorithmic methods can help address population bias. If the biases in medical imaging stem from images sourced from various centers and different device brands, a viable solution could be using disentangled representation methods for style transfer. A model that is exclusively trained on data gathered from a hospital predominantly serving white patients may not perform as effectively for other races. In such cases, the use of adversarial learning to forget demographic attributes from the model could help it focus more on disease characteristics and promote fairness.

Insufficient data, one of the potential causes of representation bias, is particularly acute in domains like biomedicine. Common tasks in CV or NLP can utilize untrained annotators, as models in the general domain aim to replicate abilities that most individuals possess. However, in biomedicine, data annotation often necessitates the involvement of highly skilled clinicians and physicians, which can be time-consuming and sometimes challenging to achieve inter-annotator consensus due to varying standards across different centers. Distributional techniques can help mitigate these issues by generating controlled synthetic data to enhance and diversify the dataset, or by facilitating collaboration between different centers by utilizing data under federated learning protocols.

This is not to imply that certain techniques are solely capable of addressing specific bias sources. Instead, each approach possesses its own merits and limitations, and in practice, multiple sources of bias can happen at the same time. The purpose of this review is to discuss the potential of different methods and provide a possibility for future research.

AI research in biomedicine has not been adequately tested for bias. Further, studies may have a limited sample size and only be validated within a single institution or region, potentially restricting their external validity when applied to other populations or when tested by operators outside of the original study.[1,2] A recent review that examined the design and the risk of bias in medical imaging found that only 54.3% of the studies surveyed provided validation with a different geographical dataset.[92] In addition, large biomedical datasets often do not provide population characteristics. This lack of population characteristics in large datasets can hinder the ability to conclude the generalizability of the results to other populations, potentially introducing bias toward specific demographics.[1]

As AI becomes more prevalent, it is important to develop ethical guidelines to ensure that these technologies are used responsibly. The weight is not only on the technical side but also in the phases of system deployment, such as data collection, and on developers, system integrators, and operators.[93] One way to develop ethical guidelines is to provide physicians who use AI systems with education on the construction of these systems, the datasets they are based on, and their limitations, which can help to ensure that physicians are aware of the potential risks and benefits of using these technologies in their practice.[94] Incorporating input from patients and communities through community advisory boards and patient advisory panels can also help ensure that AI systems are fair and representative of diverse backgrounds.[95]

## 5. CONCLUSION

As AI systems continue to be successfully implemented in various biomedicine applications, there is a growing need for research on biases in these systems. In the literature reviewed, only a small proportion of studies address algorithmic methods for addressing bias, with the majority focusing on policies.

Our review systematically covered literature only from January 2018 to December 2022. Although some works prior to this period are included, there might be more related works that are not included in this review. Our search was performed on PubMed, and other relevant articles that are not listed in the PubMed database may not be included. Our review does not cover topics related to bias metrics. There has been a lot of research in regard to designing different bias metrics, and reviews such as Xu et al. contain more details than this review.[96]

Although some debiasing methods have been applied in the biomedicine domain, there are other methods that have been used in general domains that may also be applicable in biomedicine. Different methodologies can be applied to different sources of bias, and identifying the exact source of bias is also crucial to addressing fairness in biomedicine. In addition to technical considerations, factors such as dataset diversity, community engagement, and interpretability can significantly impact the performance and utility of AI systems in biomedicine and should be carefully considered in any analysis or implementation.

Acknowledgment: This work is supported by the NIH Intramural Research Program, National Library of Medicine.